\documentclass[10pt,twocolumn,letterpaper]{article}

\usepackage{authblk}
\usepackage{wacv}              

\usepackage{times}
\usepackage{epsfig}
\usepackage{graphicx}
\usepackage{amsmath}
\usepackage{amssymb}
\usepackage{booktabs}
\usepackage{setspace}

\usepackage{nicefrac} 
\usepackage{multirow}
\usepackage{bm}
\usepackage{bbm}
\usepackage{booktabs}
\usepackage{kotex}
\usepackage{comment}
\usepackage{dsfont}
\usepackage[dvipsnames]{xcolor}
\usepackage{soul}
\usepackage[ruled,vlined]{algorithm2e}

\usepackage{mathtools}
\usepackage{xspace}

\newcommand{\mba}{\mathbf{a}}

\newcommand{\mbm}{\mathbf{m}}

\newcommand{\mbs}{\mathbf{s}}

\newcommand{\mbx}{\mathbf{x}}
\newcommand{\mby}{\mathbf{y}}

\newcommand{\mbF}{\mathbf{F}}

\newcommand{\mbH}{\mathbf{H}}

\newcommand{\mbM}{\mathbf{M}}

\newcommand{\mbT}{\mathbf{T}}

\newcommand{\ignore}[1]{}

\makeatletter
\DeclareRobustCommand\onedot{\futurelet\@let@token\@onedot}
\def\@onedot{\ifx\@let@token.\else.\null\fi\xspace}

\def\etal{{et al}\onedot}
\makeatother

\usepackage[pagebackref,breaklinks,colorlinks]{hyperref}

\usepackage[capitalize]{cleveref}
\crefname{section}{Sec.}{Secs.}
\Crefname{section}{Section}{Sections}
\Crefname{table}{Table}{Tables}
\crefname{table}{Tab.}{Tabs.}

\def\wacvPaperID{21} 
\def\confName{WACV}
\def\confYear{2024}

\begin{document}

\title{Leveraging 2D Masked Reconstruction \\for Domain Adaptation of 3D Pose Estimation}


\author{Hansoo Park\textsuperscript{1} \qquad
Chanwoo Kim\textsuperscript{1} \qquad
Jihyeon Kim\textsuperscript{1} \qquad
Hoseong Cho\textsuperscript{1} \qquad \\
Nhat Nguyen Bao Truong\textsuperscript{1} \qquad
Taehwan Kim \textsuperscript{1} \qquad
Seungryul Baek\textsuperscript{1*}}

\affil{\textsuperscript{1}UNIST, South Korea}

\maketitle

\begin{abstract}
   RGB-based 3D pose estimation methods have been successful with the development of deep learning and the emergence of high-quality 3D pose datasets. However, most existing methods do not operate well for testing images whose distribution is far from that of training data. However, most existing methods do not operate well for testing images whose distribution is far from that of training data. This problem might be alleviated by involving diverse data during training, however it is non-trivial to collect such diverse data with corresponding labels (i.e. 3D pose). In this paper, we introduced an unsupervised domain adaptation framework for 3D pose estimation that utilizes the unlabeled data in addition to labeled data via masked image modeling (MIM) framework. Foreground-centric reconstruction and attention regularization are further proposed to increase the effectiveness of unlabeled data usage. Experiments are conducted on the various datasets in human and hand pose estimation tasks, especially using the cross-domain scenario. We demonstrated the effectiveness of ours by achieving the state-of-the-art accuracy on all datasets.
\end{abstract}

\section{Introduction}
\label{sec:intro}

\begin{figure}[t]
\centering
\includegraphics[width=\linewidth]{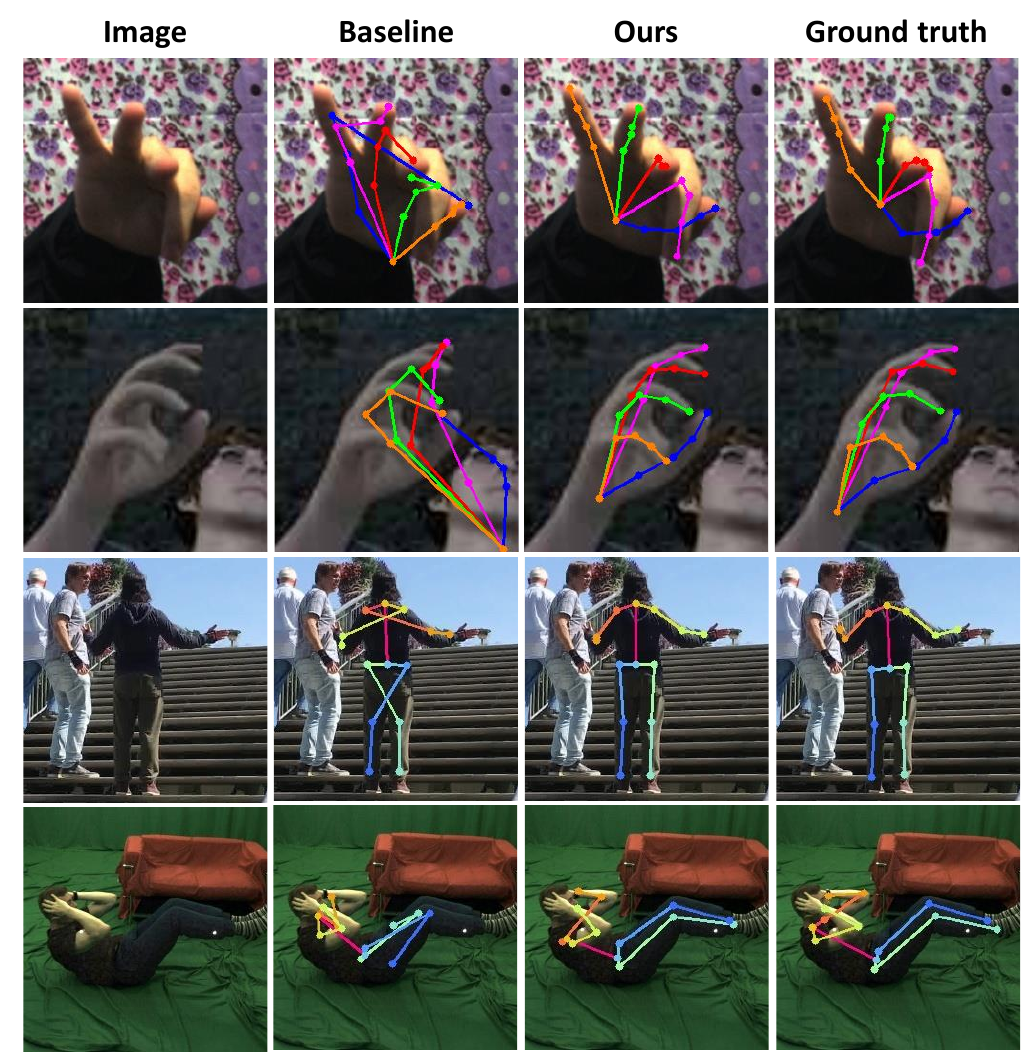}
\vspace{-2em}
\caption{Example results of pose estimation on different domains from training dataset. The second
and third columns denote the result of the baseline and our framework, respectively. The baseline denotes the
pose estimator trained with source domain dataset in a supervised manner. The baseline provides a low quality of poses on different domains, while ours provides more accurate results.}
\vspace{-1em}
\label{fig:fig1}
\end{figure}

Pose estimation from RGB images in 3D space has become an increasingly important research area due to various applications, such as Augmented Reality (AR), Virtual Reality (VR), and Human-Computer Interaction (HCI). In recent years, many researches~\cite{yang2017stacked, zimmerman_iccv2017, iqbal2018hand, baek2018augmented, ge2018hand, ge2018point, moon2018v2v, baek2019pushing, wang2020deep, lin2021end, lin2021mesh, cha2021towards, mao2021tfpose, yang2021transpose, li2021pose, zheng20213d, hampali2022keypoint, shi2022end, cha2022multi, xu2022vitpose} have been published in the computer vision field, aiming to improve the performance of pose estimation. These progresses can be attributed to the development of deep learning~\cite{krizhevsky2017imagenet, he2016deep, wang2020deep, dosovitskiy2020image} and availability of high-quality datasets~\cite{zimmerman_iccv2017, mueller2017stb, joo2018total, mueller2018ganerated, zimmermann2019freihand, ionescu2013human3, von2018recovering, mehta2017monocular, varol2017learning, johnson2010clustered, mehta2018single, lin2014microsoft, andriluka20142d, patel2021agora}, which have been made public by researchers. Despite these successful advances, most methods have suffered from the issue that model could not provide well-generalized results in different domains due to the domain gap between datasets. (see Figure~\ref{fig:fig1} for examples)  

To deal with this problem, it is crucial to study the domain adaptation approaches to ensure the robustness and reliability of the models. Numerous methods~\cite{doersch2019sim2real, zhang2019unsupervised, mu2020learning, zhao2020knowledge, spurr2021peclr, yang2022tempclr, kim2022unified, ohkawa2022domain, kundu2022uncertainty, gholami2022adaptpose, chai2023global} have been proposed to tackle the aforementioned challenge. This is typically accomplished by formulating it as the Unsupervised Domain Adaptation (UDA) problem, where the model is trained with both the labeled source domain and unlabeled target domain dataset. In~\cite{doersch2019sim2real, zhang2019unsupervised, gholami2022adaptpose, chai2023global}, they addressed this issue by using less sensitive information to the domain for predicting the 3D keypoints, such as 2D keypoints, segmentation mask and optical flow. However, these approaches require additional information beyond the labels and its performance can be dependent on the robustness of the module to estimate relevant information. In~\cite{mu2020learning, zhao2020knowledge, kim2022unified, ohkawa2022domain, kundu2022uncertainty}, they proposed the self-training strategy using the 3D pose constraint obtained from the output of network on the target domain, instead of ground-truth. However, there are certain limitations associated with these methods: 1) the performance can be highly dependent on the capacity of the network, and 2) when the gap between source and target domains is large, domain adaptation may not work well due to the noisy prediction on the target domain. Recently, the self-supervised learning methods~\cite{spurr2021peclr, yang2022tempclr} are proposed as a promising approach for improving the generalization ability of model. They aligned the distributions of both domains with contrastive learning, while reducing the dependency on additional modules or incomplete knowledge. However, there are a few limitations in these methods: 1) the contrastive learning is not suitable to align the distribution for the pose estimation task since it focuses on capturing the global representation while the pose estimation task needs both local and global representations~\cite{li2021mst}, and 2) they are suffered from the catastrophic forgetting problem for the target domain in fine-tuning stage.

In this paper, we propose the UDA framework for 3D pose estimation based on the self-supervised learning method as~\cite{spurr2021peclr, yang2022tempclr}. We use the masked image modeling (MIM) method to reconstruct the corrupted part with an uncorrupted one in the pre-training stage, instead of contrastive learning. This encourages the encoder to capture global and local representations. Additionally, we propose the foreground-centric reconstruction term to focus on reconstructing the foreground region with the segmentation mask. Through this, we encourage the network to learn the information which is helpful for pose estimation and less sensitive to domain gap by reducing the impact of background. Afterward, we fine-tune the network with the image and label of the source domain. To mitigate the forgetting problem that arises from using only source domain data, we propose the attention regularization term to leverage attention map of the target domain data. Our contributions of this paper are summarized as follows:

\begin{itemize}
\setlength\itemsep{-0.2em}

\item We propose the unsupervised domain adaptation framework for 3D pose estimation, which consists of the pre-training and fine-tuning stage. To increase the effectiveness of representation learning, we use the masked image modeling in the pre-training stage.     
  
\item To enhance the efficiency of pre-training stage, we introduce the foreground-centric reconstruction term using the segmentation mask. We also propose the attention regularization term to mitigate the forgetting problem for the target domain during fine-tuning stage.    

\item We conduct the experiments on human and hand pose estimation tasks in cross-domain scenario and achieve state-of-the-art performances on all datasets for both tasks. Through these extensive experiments, we demonstrate the effectiveness of our framework.

\end{itemize}
\section{Related Work}
\noindent \textbf{3D Pose Estimation.} 3D pose estimation infers the location of the pre-defined keypoints from the input image or video. In recent years, 3D pose estimation has gained significant attention via the success of Convolutional Neural Network (CNN) and many CNN-based methods~\cite{zimmerman_iccv2017, iqbal2018hand, wang2020deep, cha2021towards} are emerged. Zimmermann~\etal~\cite{zimmerman_iccv2017} first proposed the deep learning-based approach for estimating the 3D hand pose from a single RGB image. Iqbal~\etal~\cite{iqbal2018hand} presented a method to tackle the challenge of depth ambiguity and improve the accuracy of hand joint localization by encoding hand joint locations with a 2.5D heatmap representation. Wang~\etal~\cite{wang2020deep} provided high spatial resolution, parallel branches, and flexible upsampling for improving the accuracy of human pose estimation. Cha~\etal~\cite{cha2021towards} proposed a self-supervised learning framework for 3D human pose and shape estimation from a single 2D image, eliminating the need for additional forms of supervision signals. Recently, many transformer~\cite{vaswani2017attention} based approaches~\cite{mao2021tfpose, yang2021transpose, li2021pose, zheng20213d, lin2021end, lin2021mesh, hampali2022keypoint, shi2022end, cha2022multi, xu2022vitpose} show promising results in achieving state-of-the-art performance on various benchmarks. \cite{mao2021tfpose, yang2021transpose, li2021pose} proposed the method utilizing the transformer to encode the image feature. Lin~\etal~\cite{lin2021end, lin2021mesh} propose the novel transformer architecture to reconstruct 3D pose and mesh vertices from a single image. Hampali~\etal~\cite{hampali2022keypoint} estimated 3D poses using an attention mechanism to explicitly disambiguate the identities of the keypoints. Shi~\etal~\cite{shi2022end} adapted the method proposed in~\cite{carion2020end} for single-stage human pose estimation. Cha~\etal~\cite{cha2022multi} introduced a novel approach for refining pose by modeling inter-person relationships in multi-person scenarios. Xu~\etal~\cite{xu2022vitpose} proposed a simple yet effective baseline model based on vision transformer~\cite{dosovitskiy2020image}. \\

\noindent \textbf{Unsupervised Domain Adaptation.} Unsupervised domain adaptation (UDA) is one of the domain adaptation methods that does not use labels of the target domain. In the pose estimation tasks, such as human and hand pose estimation, many approaches~\cite{cao2019cross, doersch2019sim2real, zhang2019unsupervised, mu2020learning, wang2020predicting, zhao2020knowledge, baek2020weakly, zhang2021learning, jiang2021regressive, kim2022unified, ohkawa2022domain, kundu2022uncertainty, gholami2022adaptpose, chai2023global} have tried to address the domain gap in UDA. Cao~\etal~\cite{cao2019cross} proposed the cross-domain adaptation method to use the discriminator for minimizing the discrepancy between unlabeled and labeled animal dataset. Doersch~\etal~\cite{doersch2019sim2real} alleviated the domain gap by using the optical flow and motion of 2D keypoints, instead of the image feature vector, for estimating the human mesh in video. Zhang \etal~\cite{zhang2019unsupervised} leveraged the multi-modal data of synthetic depth dataset to gain the knowledge of 3D human pose estimation. Mu~\etal~\cite{mu2020learning} proposed to use the temporal and spatial consistency of model output for generating the pseudo labels for unlabeled real dataset and utilize them for training the model with labeled synthetic dataset. Wang~\etal~\cite{wang2020predicting} proposed the framework to estimate the camera viewpoint in addition to 3D human pose for cross-dataset generalization. Zhao~\etal~\cite{zhao2020knowledge} introduced the framework to transfer the knowledge obtained from the source to the target hand pose datasets where superior modalities are in-accessible. Baek~\etal~\cite{baek2020weakly} proposed an end-to-end trainable pipeline to adapt hand-object domain to hand-only domain by generating de-occluded hand-only images with generative adversarial network and mesh render, then utilizing them for training hand pose estimator. Zhang~\etal~\cite{zhang2021learning} applied the domain adaptation/generalization problem of 3D human pose estimation into the causal representation learning. Jiang~\etal~\cite{jiang2021regressive} proposed the domain adaptation method for regression tasks. They used the additional modules, keypoint regressor and adversarial regressor, in addition to the pose estimator. Kim~\etal~\cite{kim2022unified} reduced the discrepancy of representations between the source and target domain by using input-level and output-level cues. Ohkawa~\etal~\cite{ohkawa2022domain} proposed the self-training domain adaptation method of hand pose estimation and segmentation to use the divergence of outputs from two teacher networks for calculating the confidence of pseudo label. Kundu~\etal~\cite{kundu2022uncertainty} calculated the prediction uncertainty with the disagreement between outputs of two heads and utilized it for domain adaptation in 3D human pose estimation. Gholami~\etal~\cite{gholami2022adaptpose} generated the synthetic motions for cross-domain generalization. Chai~\etal~\cite{chai2023global} proposed the data augmentation based domain adaptation method for 3D human pose estimation utilizing the 3D pose of source dataset and the 2D pose of target dataset. Most  aforementioned methods are dependent on the additional modules or incomplete knowledge for target domain. In this paper, we propose the UDA framework for 3D pose estimation based on the self-supervised learning method to reduce the dependency on the above elements.

\begin{figure*}[t]
\centering
\includegraphics[width=\linewidth]{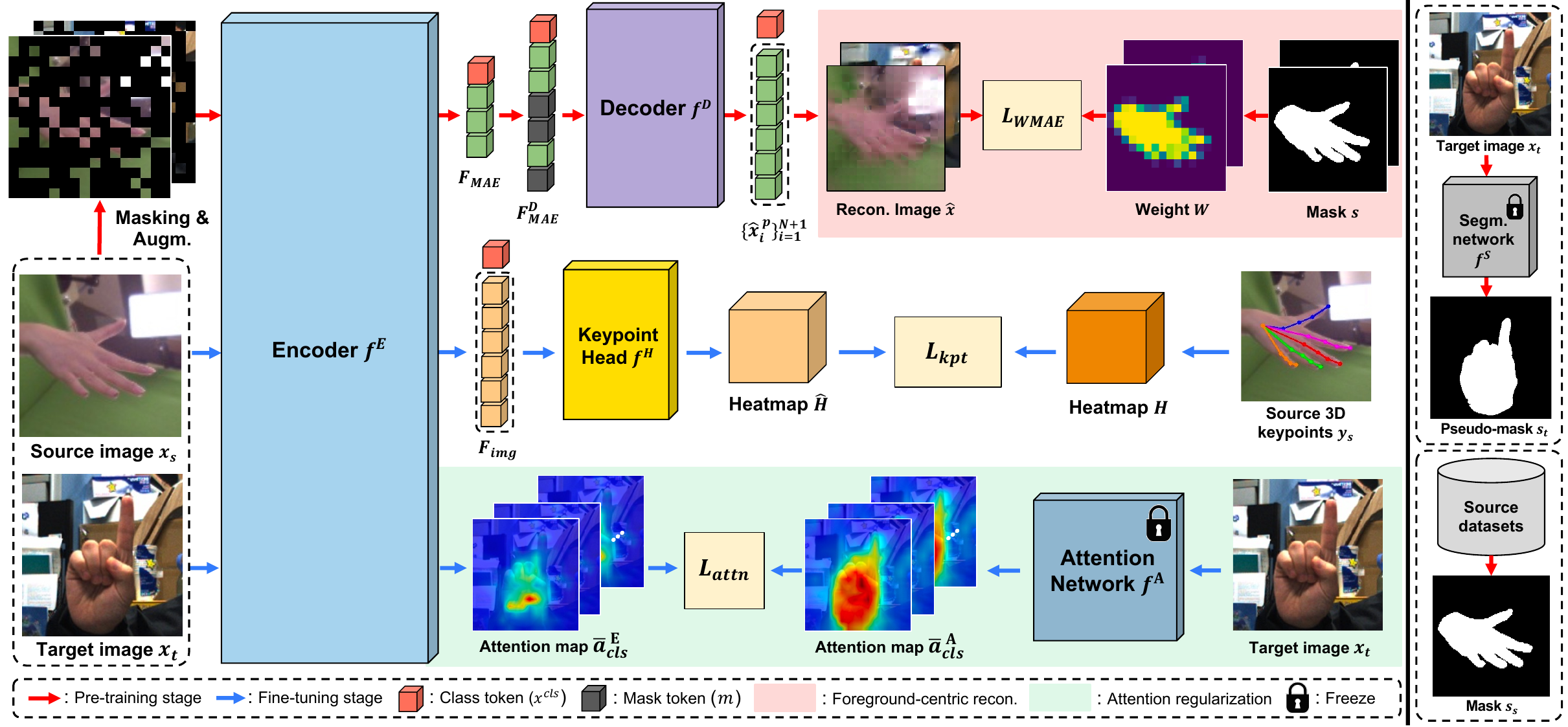}
\vspace{-0.7em}
\caption{Schematic diagram of the overall framework. Our framework consists of two stages. In pre-training stage, we perform augmentation and masking on the input image. The encoder $f^\text{E}$ extracts the latent representation $\mathbf{F}_\text{MAE}$ from the image. Then, the decoder $f^\text{D}$ reconstructs the image $\hat{\mbx}$ with the $\mathbf{F}^\text{D}_\text{MAE}$. The reconstructed image $\hat{\mbx}$ is used to compute the loss $L_\text{WMAE}$ with the corresponding image $\mbx$ and segmentation mask $s$. In fine-tuning stage, the encoder $f^\text{E}$ extracts the latent representation $\mathbf{F}_\text{img}$ from the source image $\mbx_\text{s}$. Then, the keypoint head $f^\text{H}$ estimates the heatmap $\hat{\mbH}$ with $\mathbf{F}_\text{img}$. Additionally, we obtain the attention map $\bar{\mba}^\text{E}$ and $\bar{\mba}^\text{A}$ from $f^\text{E}$ and $f^{\text{A}}$, respectively. We use the estimated heatmap $\hat{\mbH}$ and ground-truth heatmap $\mbH$ for calculating the $L_\text{kpt}$ and attention map, $\bar{\mba}^\text{E}$ and $\bar{\mba}^\text{A}$ for $L_\text{attn}$.}
\label{fig:fig2}
\end{figure*}

\section{Method}
\textbf{Overview.} Our goal is to develop a 3D pose estimator that can accurately and robustly estimate the 3D skeleton in both source and target domains. To achieve this, we utilize RGB images $X = [X_S, X_T]$ from both domains and label $Y_S$ from the source domain, which is a general setting in the unsupervised domain adaptation task. Additionally, we incorporate segmentation masks $S = [S_S, S_T]$ from both source and target domains and unconstrained dataset $X_C$, which contains images unrelated to the pose estimation task, to enhance the performance of our domain adaptation approach. Our framework consists of three trainable modules: encoder $f^\text{E}$, decoder $f^\text{D}$, and keypoint head $f^\text{H}$ and two non-trainable modules: 
segmentation network $f^\text{S}$ and attention network $f^{\text{A}}$. We use a two-stage manner, pre-training and fine-tuning, which will be detailed in Sec.~\ref{sec:pretraining} and Sec.~\ref{sec:finetune}, respectively. In the first stage, we perform the self-supervised learning of the encoder and decoder using $X$, $S$ and $f^\text{S}$. In the second stage, we fine-tune the pre-trained encoder and keypoint head with $X$, $Y_S$ and $f^\text{A}$. The overall pipeline of our framework is shown in Fig.~\ref{fig:fig2}. In the remainder of this section, we will provide more details. 

\subsection{Pre-training stage.}
\label{sec:pretraining}

In this stage, we train the encoder $f^\text{E}$ and the decoder $f^\text{D}$ with two objectives: 1) align the distribution of source and target domain, 2) extract the latent representation that is useful for pose estimation task. To achieve these goals, we employ Masked Autoencoder (MAE)~\cite{he2022masked} scheme, which is one of the Masked Image Modeling (MIM) methods. The primary reason for choosing the MAE framework is that it has higher data efficiency compared to other MIM methods, attributed to its high masking ratio, so it is applicable even in situations with limited data. We additionally propose the foreground-centric reconstruction term to encourage the encoder to extract the foreground-related representation via the segmentation mask $S$. Also, we utilize the segmentation mask for augmenting the training data with the unconstrained dataset. We use the vanilla vision transformer~\cite{dosovitskiy2020image} architecture for the $f^{\text{E}}$ and $f^{\text{D}}$ as MAE. 

\noindent \textbf{Reconstructing corrupted parts.} The encoder $f^\text{E}: X \rightarrow F$ extracts the latent representation $\mathbf{F}_\text{MAE} \in F$ from the corrupted image. The encoder $f^\text{E}$ consists of the patch embedding layer and series of the transformer blocks $[\text{T}_{1}, \text{T}_{2}, ..., \text{T}_{L}]$, where $L$ is the number of blocks. 

Given an image $\mbx\in X \subset\mathbb{R}^{H\times W\times C}$, we transform it into a sequence of non-overlapped patches $\{x_{i}\}_{i=1}^{N}$ using $(P, P)$ as the patch size and embed them into tokens $\mathbf{x}^{p} = \{x_{i}^{p}\}_{i=1}^{N} \in \mathbb{R}^{N\times 768}$ through the patch embedding layer, where $N = (H/P) \times (W/P)$. We then generate the random binary mask $\{m^{b}_{i}\}_{i=1}^{N} \in \mathbb{R}^{N}$ to contain the unmasked token set $\mathbf{x}^{p}_{m}$ which is the subset of $\mathbf{x}^{p}$, where $m^{b}_{i} \in \{0, 1\}$. The $\mathbf{x}^{p}_{m} = \{x_{i}^{p} | m^{b}_{i} = 0\}_{i=1}^{N} \in \mathbb{R}^{(1-r)N\times768}$ includes tokens $x_{i}^{p}$ if $m^{b}_{i} = 0$, and the number of tokens in $\mathbf{x}^{p}_{m}$ depends on the masking ratio $r$. After that, we concatenate the subset $\mathbf{x}^{p}_{m}$ and class token $x^\text{cls} \in \mathbb{R}^{768}$, which is the learnable parameter, for inputting to the series of transformer blocks, denoted as $F_{1} = [x^\text{cls}, \mathbf{x}^{p}_{m}] \in \mathbb{R}^{((1-r)N + 1)\times768}$. Each block consists of a Layer Normalization (LN), Multi-Head Self-Attention (MHSA) layer, and Feed-Forward Network (FFN). The input of $i$-th block $F_{i}$ is updated to $F_{i+1}$ as follows:
\begin{eqnarray}
\label{eq:eq1}
     F_{i+1}^{'} = F_{i} + {\rm MHSA}({\rm LN}(F_{i})), \nonumber \\
     \quad F_{i+1} = F_{i+1}^{'} + {\rm FFN}({\rm LN}(F_{i+1}^{'}))
\end{eqnarray}
By utilizing the MHSA layer in the transformer block, each token in $F_i$ can incorporate relational information from other tokens. We use the output of the last transformer block $\mbT_{L}$ as the latent representation $\mathbf{F}_\text{MAE} = f^\text{E}(\mathbf{x}) \in \mathbb{R}^{((1-r)N + 1)\times768}$. The decoder $f^\text{D} : F \rightarrow X$ reconstructs the image patches $\hat{\mbx} \in X$ with the latent representation. It consists of a single transformer block and prediction layer. We first project the latent representation $\mathbf{F}_\text{MAE}$ and concatenate them with mask tokens $\mbM = \{\mbm_{i}\}^{rN}_{i=1} \in \mathbb{R}^{rN \times 512}$, where each mask token $\mbm_i$ is learnable parameter. We then transform them into $\mathbf{F}^\text{D}_\text{MAE} \in F $ by rearrangement according to the order of tokens in the image. Finally, we obtain $\hat{\mbx} = \{\hat{x}_{i}^{p}\}_{i=1}^{N}$ from the decoder by excluding the class token part from the output.

\noindent \textbf{Foreground-centric reconstruction.} The main objective of MAE is to reconstruct the patches of input image assigning the same weight to all patches of the image. However, this approach is inefficient for the pose estimation task, where the information of the foreground region is more important than the background. Furthermore, the presence of significant differences in background between datasets can hinder the alignment between distributions of the source and target domains. To tackle the issue, we assign a higher weight to the patches that include the foreground compared to other patches. This encourages the encoder to focus on learning foreground-related latent representation. To calculate the weight for each patch, we utilize the foreground segmentation mask $S = [S_S, S_T]$. Since we access only the image in the target domain, we can obtain the $S_T = f^\text{S}(X_T)$ from the segmentation network $f^\text{S} : X \rightarrow S$ pre-trained with $X_S$ and $S_S$. Given the segmentation mask $\mbs\in S \subset\mathbb{R}^{H\times W\times 1}$, we first compute the pixel ratio of foreground class $\{{w^{*}_i}\}_{i=1}^{N}$ in each patch by applying a 2D average pooling operation of kernel size $(P, P)$ and stride $P$ to the segmentation mask $\mbs$. The high value $w^{*}_i$ indicates that the $i$-th patch contains a large number of foreground pixels. We then transform $\{{w^{*}_i}\}_{i=1}^{N}$ to $\{{w_i}\}_{i=1}^{N}$ as follows: 
\begin{equation}
\label{eq:eq2}
    {w}_{i} = N\frac{\hat{{w}_{i}}}{\sum_{k=1}^{N} \hat{{w}_{k}}},~~\text{where} ~~ \hat{{w}_{i}} = \text{exp}(\alpha(w^{*}_{i} - 0.5))
\end{equation}
where $\text{exp}$ denotes the exponential function and $\alpha$ is the hyper-parameter. Finally, we use $W = \{{w_i}\}_{i=1}^{N}$ as the weight values of patches.

\noindent \textbf{Data augmentation.} To enhance the data diversity, we augment the input image $X$ with segmentation mask $S$ and unconstrained dataset $X_C$. We change the background region of input image into the unconstrained image $\mbx_\text{c} \in X_\text{C}$ with the mask $\mbs \in S$, while maintaining the foreground.

\noindent \textbf{Training.} To train the encoder $f^\text{E}$ and decoder $f^\text{D}$, we employ the weighted mean squared error as the loss function $L_\text{WMAE}$. This function computes the $l2$-loss between the ground-truth patches $\mbx = \{x_{i}\}_{i=1}^{N}$ and predicted patches $\hat{\mbx}=\{ \hat{x}^{p}_{i}\}_{i=1}^{N}$ and uses $W = \{{w_{i}}\}_{i=1}^{N}$ as weights. We only use the patches updated from mask tokens for calculating the $L_\text{WMAE}$. The loss function is as follows:
\begin{eqnarray}
\label{eq:eq3}
L_{\text{WMAE}}(f^\text{E}, f^\text{D}) = \frac{1}{N}\sum_{i=1}^{N}{w}_{i}m^{b}_{i}\|\hat{x}_{i}^{p} - {x}_{i}\|^{2}_2
\end{eqnarray}

\begin{table*}[!t]
    \centering
    {\begin{footnotesize}
    \caption{Quantitative results of 3D hand pose estimation in source only and domain adaptation setting. The experiments are conducted on FreiHAND~\cite{zimmermann2019freihand}, STB~\cite{mueller2017stb}, RHD~\cite{zimmerman_iccv2017}, Panoptic (PAN)~\cite{joo2018total} and Ganerated (GAN)~\cite{mueller2018ganerated}. The best is bold-faced.}
    \resizebox{\textwidth}{!}{
    \begin{tabular}{|c|c|cc|cc|cc|cc|}
    \hline
        \multirow{2}{*}{\begin{tabular}[c]{@{}c@{}}Learning\\ Category\end{tabular}} & \multirow{2}{*}{Methods} & \multicolumn{2}{c|}{FreiHAND $\rightarrow$ STB} & \multicolumn{2}{c|}{FreiHAND $\rightarrow$ RHD} & \multicolumn{2}{c|}{FreiHAND $\rightarrow$ PAN} & \multicolumn{2}{c|}{FreiHAND $\rightarrow$ GAN} \\ \cline{3-10} 
         &  & \multicolumn{2}{c|}{EPE $\downarrow$} & \multicolumn{2}{c|}{EPE $\downarrow$} & \multicolumn{2}{c|}{EPE $\downarrow$} & \multicolumn{2}{c|}{EPE $\downarrow$} \\ \hline
        \multirow{4}{*}{\begin{tabular}[c]{@{}c@{}}Source Only\\ (FreiHand)\end{tabular}} & Zhang~\etal~\cite{zhang2021learning} & \multicolumn{2}{c|}{36.1} & \multicolumn{2}{c|}{48.3} & \multicolumn{2}{c|}{35.6} & \multicolumn{2}{c|}{59.4} \\  \cline{2-10} 
         & Ours (Scratch) & \multicolumn{2}{c|}{27.0} & \multicolumn{2}{c|}{46.3} & \multicolumn{2}{c|}{34.6} & \multicolumn{2}{c|}{67.3} \\ 
         & Ours (MAE)~\cite{he2022masked} & \multicolumn{2}{c|}{24.7} & \multicolumn{2}{c|}{41.7} & \multicolumn{2}{c|}{31.9} & \multicolumn{2}{c|}{58.1} \\ 
         & Ours w/o $X_C$ & \multicolumn{2}{c|}{22.7} & \multicolumn{2}{c|}{38.6} & \multicolumn{2}{c|}{30.1} & \multicolumn{2}{c|}{53.0} \\
         & Ours & \multicolumn{2}{c|}{\textbf{22.0}} & \multicolumn{2}{c|}{\textbf{37.1}} & \multicolumn{2}{c|}{\textbf{29.1}} & \multicolumn{2}{c|}{\textbf{52.3}} \\ \hline
         
        \multirow{6}{*}{\begin{tabular}[c]{@{}c@{}}Domain\\ Adaptation\end{tabular}} & DDC~\cite{tzeng2014deep} & \multicolumn{2}{c|}{34.5} & \multicolumn{2}{c|}{44.6} & \multicolumn{2}{c|}{32.5} & \multicolumn{2}{c|}{57.3} \\  
         & DAN~\cite{long2015learning} & \multicolumn{2}{c|}{32.7} & \multicolumn{2}{c|}{40.5} & \multicolumn{2}{c|}{32.1} & \multicolumn{2}{c|}{54.9} \\  
         & DANN~\cite{ganin2016domain} & \multicolumn{2}{c|}{30.9} & \multicolumn{2}{c|}{38.0} & \multicolumn{2}{c|}{31.8} & \multicolumn{2}{c|}{53.6}  \\  
         & Zhang~\etal~\cite{zhang2021learning} & \multicolumn{2}{c|}{22.4} & \multicolumn{2}{c|}{35.4} & \multicolumn{2}{c|}{22.9} & \multicolumn{2}{c|}{49.5} \\  \cline{2-10} 
         & Ours (MAE) & \multicolumn{2}{c|}{21.2} & \multicolumn{2}{c|}{34.4} & \multicolumn{2}{c|}{25.7} & \multicolumn{2}{c|}{45.9} \\ 
         & Ours w/o $X_C$ & \multicolumn{2}{c|}{19.7} & \multicolumn{2}{c|}{29.1} & \multicolumn{2}{c|}{22.0} & \multicolumn{2}{c|}{35.2} \\
         & Ours & \multicolumn{2}{c|}{\textbf{19.2}} & \multicolumn{2}{c|}{\textbf{28.0}} & \multicolumn{2}{c|}{\textbf{21.2}} & \multicolumn{2}{c|}{\textbf{33.2}} \\ \hline
    \end{tabular}
    \label{tab:DA_qualitative_hand}}
    \end{footnotesize}}
\end{table*}

\begin{figure*}[!t]
\centering
\includegraphics[width=0.98\linewidth]{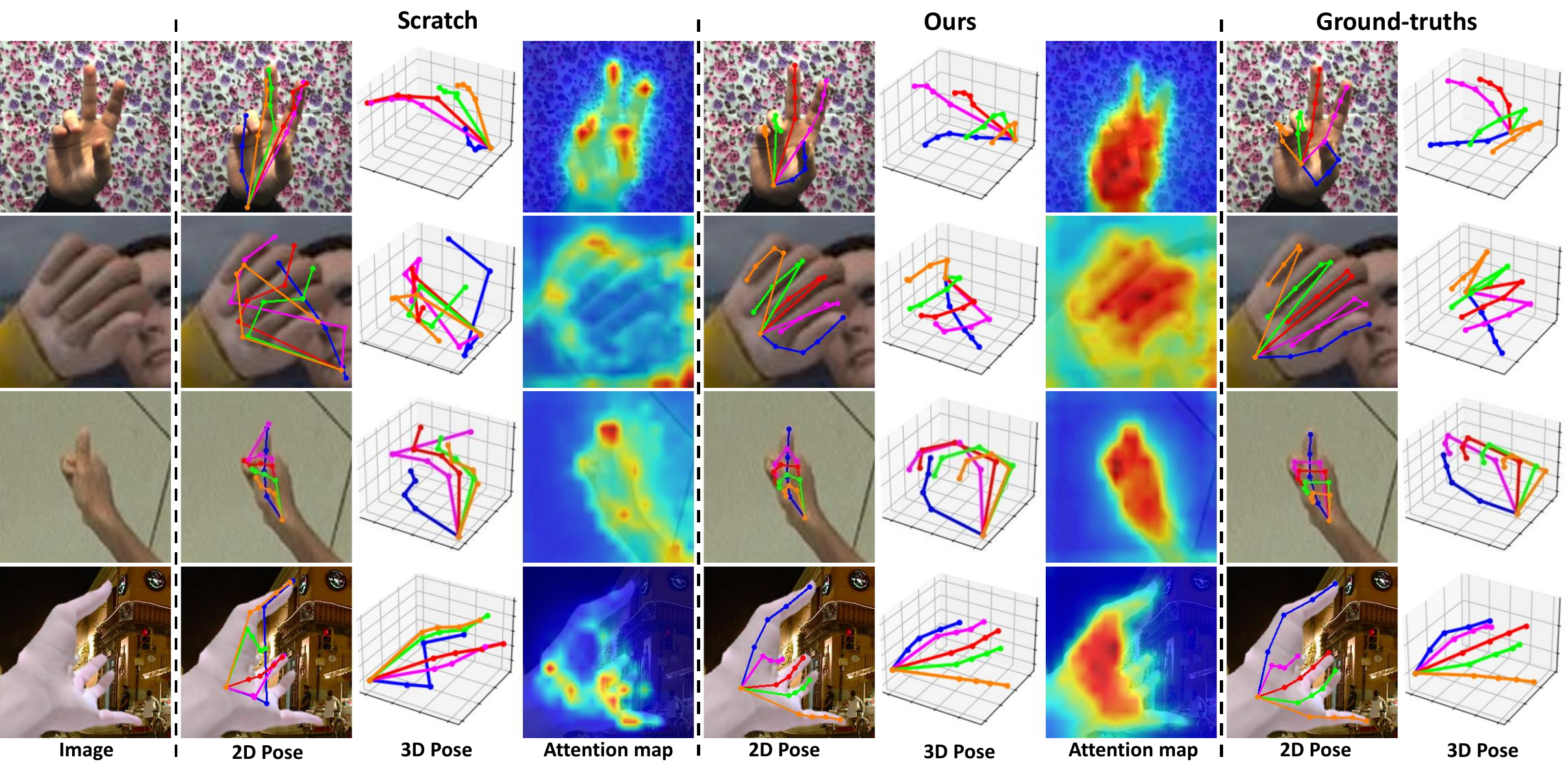}
\caption{Qualitative comparisons of scratch and ours on (first row) STB~\cite{mueller2017stb}, (second row) RHD~\cite{zimmerman_iccv2017}, (third row) Panoptic (PAN)~\cite{joo2018total}, and (last row) Ganerated (GAN)~\cite{mueller2018ganerated} datasets. We involve the results of 2D pose, 3D pose and attention map for scratch and ours, respectively.}
\label{fig:fig3}
\end{figure*}

\subsection{Fine-tuning stage.}
\label{sec:finetune}

In this stage, we fine-tune the pre-trained encoder $f^\text{E}$ and keypoint head $f^\text{H}$ with $X_S$ and $Y_S$. We follow the architecture of~\cite{xu2022vitpose}, which uses the vision transformer as the encoder and estimates the keypoint as the heatmap. Additionally, we use $X_T$ and attention network $f^{\text{A}}$ to preserve information specific to the target domain. The network $f^{\text{A}}$ has the same architecture and weight with pre-trained $f^\text{E}$, but is not updated during the fine-tuning stage.




\noindent \textbf{3D pose estimation.} We extract the latent representation $\mbF_\text{img} = f^\text{E}(\mbx_s) \in \mathbb{R}^{(N+1)\times768}$ from the input image of source domain $\mbx_s \in X_S$ and the class token $x^\text{cls}$ with the encoder $f^\text{E}$ in the same way as in pre-training stage, but with all the tokens. Then, we exclude the class token part from $\mbF_\text{img}$ and transform it to $\mathbb{R}^{\sqrt{N} \times \sqrt{N} \times 768}$. Finally, we obtain the 3D heatmaps $\hat{\mbH} = f^\text{H}(\mbF_\text{img}) \in \mathbb{R}^{K \times 4\sqrt{N} \times 4\sqrt{N} \times 4\sqrt{N}}$ from the keypoint head $f^\text{H}:F \rightarrow H$ as follows:
\begin{eqnarray}
\label{eq:eq4}
\hat{\mbH} = \text{Conv}(\text{Deconv}(\text{Deconv}(\mbF_\text{img}))
\end{eqnarray}
where K denotes the number of keypoints. Additionally, $\text{Deconv}$ denotes the deconvolution layer with $4\times 4$ kernels and $2$ stride, and $\text{Conv}$ denotes $1\times 1$ convolution layer. 


\noindent \textbf{Attention regularization.} Through the pre-training stage, the encoder $f^\text{E}$ has higher attention with input tokens of the foreground region than the background, and this attention is maintained for the source domain data after the fine-tuning stage. However, it is not maintained in the target domain since the network $f^\text{E}$ is trained for pose estimation using only the source domain dataset during the fine-tuning stage. To alleviate the forgetting problem of target domain information, we propose the attention regularization term to maintain the attention for target domain data by utilizing the attention map of $X_T$ obtained from the attention network $f^\text{A}$. Given an image $\mbx_t \in X_T$, we pass it to the patch embedding layer and series of transformer blocks in $f^\text{A}$. We then obtain the attention matrix $A_i = [a^i_1, a^i_2, ..., a^i_\text{cls}]$ from MHSA layer of $i$-th transformer block, where $a^{i}_{j} \in \mathbb{R}^{N}$ denotes the attention vector of $j$-th token excluding the $j$-th element. Since the class token is better able to consider the global relationship rather than other tokens~\cite{caron2021emerging}, we only use the $a^i_\text{cls}$ in the attention matrix of i-th block. Therefore, we obtain the attention map $\bar{\mba}^\text{A}_\text{cls} = \{a^1_\text{cls}, a^2_\text{cls}, ..., a^L_\text{cls}\}$ consisting of the attention vector in each block. We compare $\bar{\mba}^\text{A}_\text{cls}$ to $\bar{\mba}^\text{E}_\text{cls}$, which is the attention map of encoder $f^\text{E}$. Through this approach, we can encourage the encoder $f^\text{E}$ to maintain the foreground-related information of the target domain.

\noindent \textbf{Training.} To train the network $f^\text{E}$ and $f^\text{H}$, we use the $L_\text{kpt}$ that computes $l2$-loss between the predicted heatmaps $\hat{\mbH}$ and ground-truth heatmaps $\mbH$ as follows:
\begin{eqnarray}
\label{eq:eq6}
L_{\text{kpt}}(f^\text{E}, f^\text{H}) = \left \|  \hat{\mbH} - \mbH \right \|^{2}_{2}
\end{eqnarray}
To obtain the ground-truth heatmaps of 3D keypoints $\mby_s \in Y_S \subset\mathbb{R}^{K \times 3}$, we use the Gaussian blob. Additionally, we use the $L_{\text{attn}}$ to regularize the encoder $f^\text{E}$. The loss function is as follows:
\begin{eqnarray}
\label{eq:eq7}
L_{\text{attn}}(f^\text{E}) = \left \|  \bar{\mba}^{E}_\text{cls} - \bar{\mba}^\text{A}_\text{cls} \right \|_{1}
\end{eqnarray}
Finally, we use the linear combination of two losses to train the network $f^\text{E}$ and $f^\text{H}$ as follows:
\begin{eqnarray}
\label{eq:eq5}
L_\text{HPE}(f^\text{E}, f^\text{H}) =  L_{\text{kpt}}(f^\text{E}, f^\text{H}) + \lambda_\text{attn} L_{\text{attn}}(f^\text{E})
\end{eqnarray}
We will provide the details of network architecture and implementation in the supplementary material.

\section{Experiment}
\label{sec:experiment}
To demonstrate the effectiveness of our framework, we perform two kinds of experimental settings: source-only and domain adaptation. In the source-only setting, we train the framework using only source domain data. We do not use the $L_\text{attn}$ and $f^\text{S}$ requiring the target domain data for training. In the domain adaptation setting, we use both source and target domain dataset for training but do not access the label of target domain dataset. The evaluation is conducted on the target domain datasets in both settings.



\subsection{Datasets and Evaluation metrics}
\noindent \textbf{Hand pose estimation.} We performed experiments for hand pose estimation with five datasets: FreiHAND~\cite{zimmermann2019freihand}, STB~\cite{mueller2017stb}, RHD~\cite{zimmerman_iccv2017}, Panoptic (PAN)~\cite{joo2018total} and Ganerated (GAN)~\cite{mueller2018ganerated}, which contain RGB images and 3D hand pose labels. The details can be found in the supplementary material. For the evaluation metrics, we use mean End-Point-Error (EPE). EPE is the average Euclidean distance between estimated and ground-truth keypoints. We used the millimeters (mm) as the metric unit.

\begin{table*}[!t]
    \centering
    {\begin{footnotesize}
    \caption{Quantitative results of 3D human pose estimation in source only and domain adaptation setting. The experiment is conducted on Human3.6M(H3.6M)~\cite{ionescu2013human3}, 3DPW~\cite{von2018recovering}, MPI-INF-3DHP (3DHP)~\cite{mehta2017monocular} and SURREAL~\cite{varol2017learning}. The best is bold-faced.}
    \resizebox{\textwidth}{!}{
    \begin{tabular}{|c|c|cc|cc|cc|}
    \hline
        \multirow{2}{*}{\begin{tabular}[c]{@{}c@{}}Learning\\ Category\end{tabular}} & \multirow{2}{*}{Methods} & \multicolumn{2}{c|}{H3.6M $\rightarrow$ 3DPW} & \multicolumn{2}{c|}{H3.6M $\rightarrow$ 3DHP} & \multicolumn{2}{c|}{H3.6M $\rightarrow$ SURREAL} \\ \cline{3-8} 
         &  & \multicolumn{1}{c|}{MPJPE $\downarrow$} & PA-MPJPE $\downarrow$ & \multicolumn{1}{c|}{MPJPE $\downarrow$} & PA-MPJPE $\downarrow$ & \multicolumn{1}{c|}{MPJPE $\downarrow$} & PA-MPJPE $\downarrow$ \\ \hline
        \multirow{4}{*}{\begin{tabular}[c]{@{}c@{}}Source Only\\ (H3.6M)\end{tabular}} & Zhang~\etal~\cite{zhang2021learning} & \multicolumn{1}{c|}{118.7} & 78.0 & \multicolumn{1}{c|}{121.8} & 98.5 & \multicolumn{1}{c|}{128.6} & 86.5\\ \cline{2-8}
         & Ours (Scratch) & \multicolumn{1}{c|}{87.2} & 60.8 & \multicolumn{1}{c|}{120.5} & 92.3 & \multicolumn{1}{c|}{113.0} & 74.7 \\ 
         & Ours (MAE)~\cite{he2022masked} & \multicolumn{1}{c|}{83.4} & 58.5 & \multicolumn{1}{c|}{113.5} & 87.7 & \multicolumn{1}{c|}{106.7} & 72.0\\ 
         & Ours w/o $X_C$ & \multicolumn{1}{c|}{82.0} & 57.2 & \multicolumn{1}{c|}{109.2} & 88.1 & \multicolumn{1}{c|}{105.0} & 71.4 \\ 
         & Ours & \multicolumn{1}{c|}{\textbf{80.0}} & \textbf{56.8} & \multicolumn{1}{c|}{\textbf{107.0}} & \textbf{83.1} & \multicolumn{1}{c|}{\textbf{103.2}} & \textbf{68.6} \\ \hline
        \multirow{8}{*}{\begin{tabular}[c]{@{}c@{}}Domain\\ Adaptation\end{tabular}} & DDC~\cite{tzeng2014deep} & \multicolumn{1}{c|}{110.4} & 75.3 & \multicolumn{1}{c|}{115.6} & 91.5 & \multicolumn{1}{c|}{117.5} & 80.1 \\  
         & DAN~\cite{long2015learning} & \multicolumn{1}{c|}{107.5} & 73.2 & \multicolumn{1}{c|}{109.5} & 89.2 & \multicolumn{1}{c|}{114.2} & 78.4\\ 
         & DANN~\cite{ganin2016domain} & \multicolumn{1}{c|}{106.3} & 71.1 & \multicolumn{1}{c|}{107.9} & 88.0 & \multicolumn{1}{c|}{113.6} & 77.2\\  
         & ISO~\cite{zhang2020inference} & \multicolumn{1}{c|}{-} & 70.8 & \multicolumn{1}{c|}{-} & 75.8 & \multicolumn{1}{c|}{-} & -\\  
         & Zhang~\etal~\cite{zhang2021learning} & \multicolumn{1}{c|}{94.7} & 63.9 & \multicolumn{1}{c|}{99.3} & 81.5 & \multicolumn{1}{c|}{103.3} & 69.1\\ 

         & Kundu~\etal~\cite{kundu2022uncertainty} & \multicolumn{1}{c|}{91.9} & 62.1 & \multicolumn{1}{c|}{96.2} & 78.6 & \multicolumn{1}{c|}{99.6} & 67.2\\  \cline{2-8} 
         
         & Ours (MAE) & \multicolumn{1}{c|}{83.0} & 58.4 & \multicolumn{1}{c|}{101.3} & 80.9 & \multicolumn{1}{c|}{94.1} & 62.7 \\ 
         & Ours w/o $X_C$ & \multicolumn{1}{c|}{80.0} & 55.6 & \multicolumn{1}{c|}{94.6} & 75.0 & \multicolumn{1}{c|}{86.5} & 58.0 \\ 
         & Ours & \multicolumn{1}{c|}{\textbf{78.1}} & \textbf{54.9} & \multicolumn{1}{c|}{\textbf{93.6}} & \textbf{74.8} & \multicolumn{1}{c|}{\textbf{83.1}} & \textbf{55.0}\\ \hline
        \end{tabular}
        \label{tab:DA_qualitative_human}}
        \end{footnotesize}}
\end{table*}

\begin{figure*}[!t]
\centering
\vspace{-0.5em}
\includegraphics[width=\linewidth]{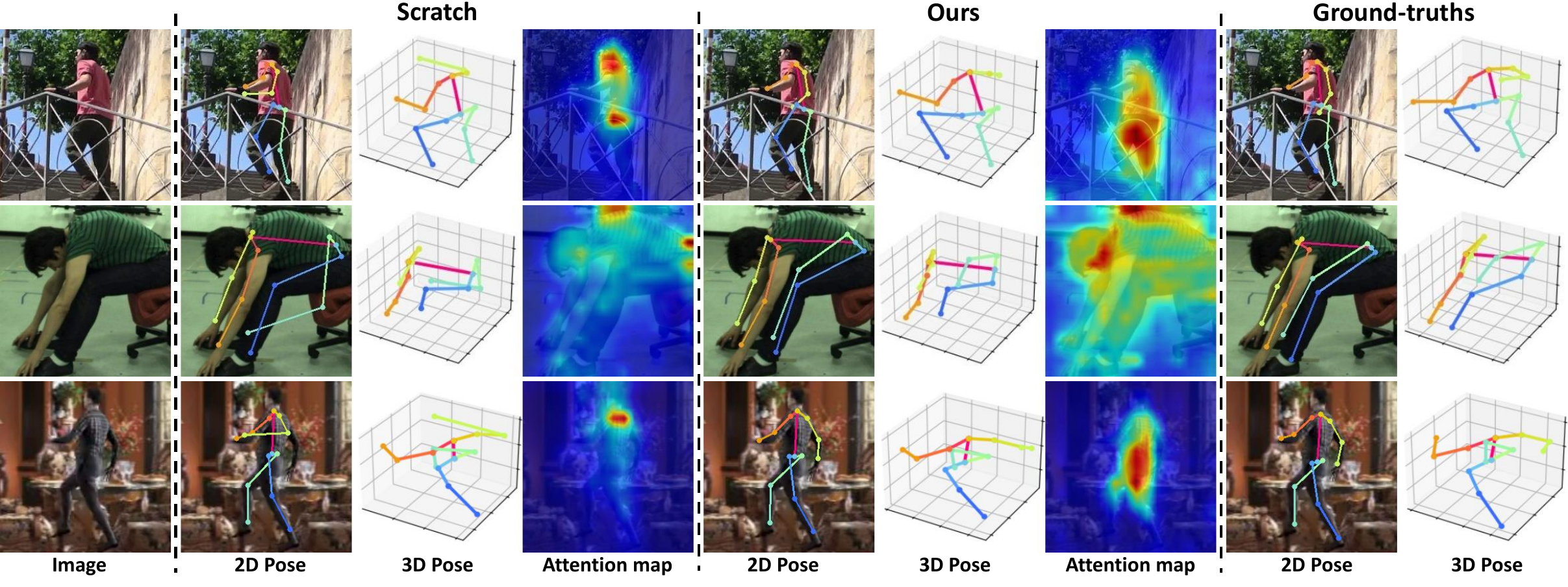}
\caption{Qualitative comparisons of scratch and ours on (first row) 3DPW~\cite{von2018recovering}, (second row) MPI-INF-3DHP (3DHP)~\cite{mehta2017monocular}, and (last row) SURREAL~\cite{varol2017learning} datasets. We involve the results of 2D pose, 3D pose and attention map for scratch and ours, respectively.}
\label{fig:fig4}
\end{figure*}

\noindent \textbf{Human pose estimation.} We performed experiments for human pose estimation with four datasets: Human3.6M~\cite{ionescu2013human3}, 3DPW~\cite{von2018recovering}, MPI-INF-3DHP(3DHP)~\cite{mehta2017monocular} and SURREAL~\cite{varol2017learning} which contain RGB images and 3D human pose labels. The details can be found in the supplementary material. For the evaluation metrics, we use the Mean Per Joint Position Error (MPJPE) and Procrustes Aligned Mean Per Joint Position Error (PA-MPJPE). MPJPE computes the Euclidean distance between the predicted and ground-truth pose in the root-joint (Pelvis) aligned. PA-MPJPE calculates the Euclidean distance between the predicted and ground-truth pose after rigidly aligning the predicted pose to ground-truth via Procrustes Analysis~\cite{gower1975generalized}. We used the millimeters (mm) as the metric unit.

\noindent \textbf{Unconstrained dataset.} We utilized the ImageNet~\cite{krizhevsky2017imagenet} dataset for the unconstrained dataset, which contains the various categories.

\begin{figure*}[t]
\centering
\includegraphics[width=\linewidth]{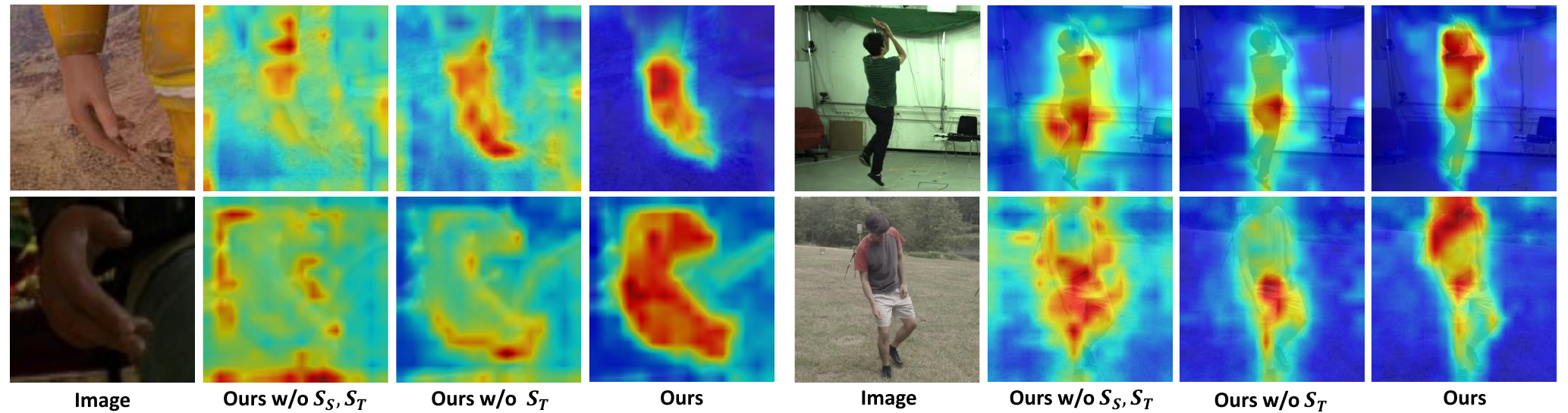}
\caption{Qualitative comparisons of attention map for target domain data $X_T$.}
\label{fig:fig5}
\end{figure*}

\begin{table*}[t]
    \vspace{-0.7em}
    \mbox{}\hfill
    \begin{minipage}[t]{.5\linewidth}
    \centering
    \footnotesize
    \caption{Ablation study of design choices on hand and human pose estimation task. Each column represents the results of each task. The best is bold-faced.}
    \scriptsize
    \resizebox{0.8\textwidth}{!}{
    \begin{tabular}{lcc}
    \hline
    \multirow{2}{*}{\textbf{Method}} & EPE($\downarrow$) & MPJPE($\downarrow$) \\
                                     & (Hand) & (Human) \\ \hline
    Ours w/o FCR, AR & 34.4 & 101.3 \\
    Ours w/o FCR & 34.6 & 104.0 \\ 
    Ours w/o AR & 32.0 & 98.1 \\ 
    Ours & \textbf{28.0} & \textbf{93.6} \\  \hline
    \end{tabular}
    }
    \label{tab:ablation_design_choices}
    \end{minipage}
    \hspace{0.7cm}
    \begin{minipage}[t]{0.45\linewidth}
    \centering
    \caption{Ablation study of effectiveness of segmentation mask on hand and human pose estimation task. Each column represents the results of each task. The best is bold-faced.}
    \scriptsize
    \resizebox{1\textwidth}{!}{
    \begin{tabular}{lcc}
    \hline
    \multirow{2}{*}{\textbf{Method}} & EPE($\downarrow$) & MPJPE($\downarrow$) \\
                                     & (Hand) & (Human) \\ \hline
    Ours w/o $S_S$,$S_T$ & 34.6 & 104.0 \\
    Ours w/o $S_T$ & 32.3 & 100.6 \\ 
    Ours & \textbf{28.0} & \textbf{93.6} \\ \hline
    \end{tabular}
    }
    
    \label{tab:ablation_mask}  
    \end{minipage} 
\end{table*}

\subsection{Experimental results}
We compared our framework with existing methods that estimate the 3D pose of target domain data in unsupervised domain adaptation. Also, we measured the performance of our baselines, denoted as Ours (Scratch), Ours (MAE) and Ours w/o $X_C$ in Table~\ref{tab:DA_qualitative_hand} and~\ref{tab:DA_qualitative_human}. Ours (Scratch) denotes the model using the $L_\text{kpt}$ for training. Ours (MAE) performs both pre-training and fine-tuning stages as~\cite{he2022masked}, but not using foreground-centric reconstruction term and attention regularization term. Ours w/o $X_C$ excludes the data augmentation utilizing the unconstrained dataset $X_C$.  

\noindent \textbf{Hand pose estimation.} According to~\cite{zhang2021learning}, we used FreiHAND~\cite{zimmermann2019freihand} as source domain dataset and STB~\cite{mueller2017stb}, RHD~\cite{zimmerman_iccv2017}, PAN~\cite{joo2018total} and GAN~\cite{mueller2018ganerated} as target domain datasets. In Table~\ref{tab:DA_qualitative_hand}, we show the quantitative results. In both experimental settings, our method could achieve state-of-the-art performance on all of the target domain datasets, compared to Zhang~\etal~\cite{zhang2021learning} and our baselines. In Fig.~\ref{fig:fig3}, we show the qualitative comparisons of 2D pose, 3D pose and attention map between Ours(Scratch) and Ours. To visualize the attention map clearly, we use the mean of attention maps in $\bar{\mba}^{E}_\text{cls}$ for visualization. Our method produces much better results on the cross-domain scenario than scratch. Ours also activates the hand region of target domain data while scratch fails to do it.       

\noindent \textbf{Human pose estimation.} We used Human3.6M~\cite{ionescu2013human3} as source domain dataset and 3DPW~\cite{von2018recovering}, MPI-INF-3DHP~\cite{mehta2017monocular} and SURREAL~\cite{varol2017learning} as target domain dataset. The quantitative results are shown in Table~\ref{tab:DA_qualitative_human}. We outperformed the existing methods, Zhang~\etal~\cite{zhang2021learning} and Kundu~\etal~\cite{kundu2022uncertainty}, and baselines on both experimental settings. Fig.~\ref{fig:fig4} shows the qualitative comparisons of pose and attention map between Ours(Scratch) and Ours. Our method produces much better results than scratch.

\subsection{Ablation Studies}
We conducted various experiments for the components of the proposed method. In all ablation studies, we used FreiHAND and RHD datasets as source and target domain datasets in hand pose estimation task. In human pose estimation task, Human3.6M and MPI-INF-3DHP dataset are used as source and target domain dataset, respectively.

\noindent \textbf{Design choices.}
We assessed the several design choices of our framework to demonstrate the effectiveness of the proposed terms: 1) Foreground-centric reconstruction term (FCR) and 2) Attention regularization term (AR). We constructed three variations, `Ours w/o FCR', `Ours w/o AR' and `Ours w/o FCR, AR'. `Ours w/o FCR' denotes our framework not using the foreground-centric reconstruction term by assigning 1 to $w_{i}$ instead of values obtained from the segmentation mask in Eq.~\ref{eq:eq3}. `Ours w/o AR' does not use the attention regularization term in our framework by removing the contribution of $L_\text{attn}$ in Eq.~\ref{eq:eq5}. `Ours w/o FCR, AR' is the framework not using both terms, same as MAE~\cite{he2022masked}. Table~\ref{tab:ablation_design_choices} shows the quantitative results on both tasks. We could demonstrate that the proposed terms are effective to improve the performance of pose estimation and domain adaptation. Also, we could observe that the case of `Ours w/o FCR' shows worse performance than `Ours w/o FCR, AR'. We will provide a comprehensive explanation for this result in the subsequent ablation study. 

\noindent \textbf{Effectiveness of segmentation mask.}
To demonstrate the effectiveness of utilizing both source domain $S_S$ and target domain $S_T$ masks during the pre-training stage, we explored various scenarios for using the segmentation mask in Table~\ref{tab:ablation_mask}; `Ours w/o $S_S$, $S_T$', `Ours w/o $S_T$' and `Ours'. We assigned a value of 1 to $w_i$ in Eq.~\ref{eq:eq3}, when the segmentation mask of a certain domain is not used. `Ours w/o $S_S$, $S_T$' is same as `Ours w/o FCR' in Table~\ref{tab:ablation_design_choices} and `Ours w/o $S_T$' does not use the target domain mask $S_T$ in FCR. We can show that the results are very limited compared to `Ours'. We additionally present the visualization of attention map $\bar{\mba}^\text{A}_\text{cls}$ for target domain data $X_T$ in Fig.~\ref{fig:fig5}. The attention map of `Ours' mainly has high attention on the foreground region, while not in other cases. In particular, the attention map is severely disrupted in the `Ours w/o $S_S$, $S_T$'. We consider it as the reason for the performance drop of `Ours w/o FCR' in Table~\ref{tab:ablation_design_choices}, since it uses the low-quality attention map $\bar{\mba}^\text{A}_\text{cls}$ for the $L_\text{attn}$. Through the above experiments, we can show that the the quality of attention map is affected by using the segmentation mask. Therefore, it is necessary to use $S_S$ and $S_T$ during the pre-training stage for improving the predicted poses.

\section{Conclusion}
In this paper, we propose the unsupervised domain adaptation framework for 3D pose estimation. Our framework consists of two stages, pre-training stage and fine-tuning stage. In the pre-training stage, we perform the masked image modeling with images of source and target domain dataset and increase the efficiency of pre-training with the foreground-centric reconstruction term. Through this term, we encourage the network to learn the information which is helpful for pose estimation and domain adaptation tasks. In the fine-tuning stage, we train the network in a supervised manner with source domain dataset. To mitigate the forgetting problem for target domain, we propose the attention regularization term to utilize the attention map of target domain data. We achieved state-of-the-art performance on both hand and human pose estimation tasks in cross-domain scenarios. Moreover, we demonstrated that our proposed terms are effective in pose estimation and domain adaptation.

{\small
\bibliographystyle{ieee_fullname}
\bibliography{egbib}
}

\twocolumn[{
\begin{center}
\textbf{\Large Leveraging 2D Masked Reconstruction \\for Domain Adaptation of 3D Pose Estimation\\-Supplementary-\\}
\end{center}
\vspace{16mm}
}]

In this supplemental, we provide the overall of training procedures in our framework; the implementation detail; the summary of datasets; the qualitative comparisons on target domain datasets; and the results of additional experiment.

\section{Overall of training procedures.}
We introduce the details of network architecture for our framework in Table.~\ref{tab:pretrain_E_D},~\ref{tab:pretrain_s},~\ref{tab:finetune_E_H} and~\ref{tab:finetune_a}. We also introduce the overall of training procedures in Algorithm.~\ref{a:training_procedure}.

\begin{table}[h]   
\caption{Architecture of encoder $f^{\text{E}}$ and decoder $f^{\text{D}}$ in pre-training stage. We use the architecture of~\cite{vaswani2017attention} as Transformer block.} 
\vspace{0.5em}
\label{tab:pretrain_E_D}
\centering
\scriptsize
\resizebox{\linewidth}{!}{
\begin{tabular}{c c c}
\toprule
Layer & Operation & Dimensionality\\
\midrule
& Input: rgb image $\mbx$ & $256 \times 256 \times 3$\\
\midrule
1 & Patch embedding $(\mathbf{x}^{p})$ & $256 \times 768$ \\
2 & Masking $(\mathbf{x}^{p}_{m})$ & $64 \times 768$ \\
\midrule
3 & Class token $x^\text{cls}$ & $1 \times 768$ \\
4 & Concat(L2, L3) & $65 \times 768$ \\ 
\midrule
5 & Transformer block & $65 \times 768$ \\
6 & Transformer block & $65 \times 768$ \\
7 & Transformer block & $65 \times 768$ \\
8 & Transformer block & $65 \times 768$ \\
9 & Transformer block & $65 \times 768$ \\
10 & Transformer block & $65 \times 768$ \\
11 & Transformer block & $65 \times 768$ \\
12 & Transformer block & $65 \times 768$ \\
13 & Transformer block & $65 \times 768$ \\
14 & Transformer block & $65 \times 768$ \\
15 & Transformer block & $65 \times 768$ \\
16 & Transformer block ($\mbF_\text{MAE}$)& $65 \times 768$ \\
\midrule
17 & Linear & $65\times 512$ \\
\midrule
18 & Mask tokens $M$ & $192\times 512$\\
\midrule
19 & Concat(L17, L18) $(\mathbf{F}^\text{D}_\text{MAE})$ & $257 \times 512$ \\ 
20 & Transformer block & $257 \times 512$ \\
21 & Linear & $257 \times 768$ \\
22 & Exclude & $256 \times 768$ \\
\midrule
    & Output: L22 $(\hat{\mbx})$ & - \\
\bottomrule
\end{tabular}
}
\vspace{-1em}
\end{table}

\begin{table}[h]   
\caption{Architecture of segmentation network $f^{\text{S}}$ in pre-training stage.}
\vspace{0.5em}
\label{tab:pretrain_s}
\centering
\scriptsize
\resizebox{\linewidth}{!}{
\begin{tabular}{c c c c}
\toprule
Layer & Operation & Kernel & Dimensionality\\
\midrule
& Input: rgb image $\mbx$ & - & $256 \times 256 \times 3$\\
\midrule
1 & Conv & $7 \times 7$ & $128 \times 128 \times 64$ \\
2 & BatchNorm + ReLU & - & $128 \times 128 \times 64$ \\
\midrule
3 & Conv + ReLU & $3 \times 3$ & $128 \times 128 \times 64$ \\
4 & Conv + ReLU & $3 \times 3$ & $128 \times 128 \times 64$ \\
5 & Max Pooling & $2 \times 2$ & $128 \times 128 \times 64$ \\

6 & Conv + ReLU & $3 \times 3$ & $64 \times 64 \times 64$ \\
7 & Conv + ReLU & $3 \times 3$ & $64 \times 64 \times 64$ \\
8 & Max Pooling & $2 \times 2$ & $32 \times 32 \times 64$ \\

9 & Conv + ReLU & $3 \times 3$ & $32 \times 32 \times 64$ \\
10 & Conv + ReLU & $3 \times 3$ & $32 \times 32 \times 64$ \\
11 & Max Pooling & $2 \times 2$ & $16 \times 16 \times 64$ \\

12 & Conv + ReLU & $3 \times 3$ & $16 \times 16 \times 64$ \\
13 & Conv + ReLU & $3 \times 3$ & $16 \times 16 \times 64$ \\
14 & Max Pooling & $2 \times 2$ & $8 \times 8 \times 64$ \\

15 & Conv + ReLU & $3 \times 3$ & $8 \times 8 \times 64$ \\
16 & Conv + ReLU & $3 \times 3$ & $8 \times 8 \times 64$ \\

\midrule
17 & Upsampling & $2 \times 2$ & $16 \times 16 \times 64$ \\
18 & Concat(L16, L17) & - & $16 \times 16 \times 128$ \\
19 & Conv + ReLU & $3 \times 3$ & $16 \times 16 \times 64$ \\
20 & Conv + ReLU & $3 \times 3$ & $16 \times 16 \times 64$ \\

21 & Upsampling & $2 \times 2$ & $32 \times 32 \times 64$ \\
22 & Concat(L13, L21) & - & $32 \times 32 \times 128$ \\
23 & Conv + ReLU & $3 \times 3$ & $32 \times 32 \times 64$ \\
24 & Conv + ReLU & $3 \times 3$ & $32 \times 32 \times 64$ \\

25 & Upsampling & $2 \times 2$ & $64 \times 64 \times 64$ \\
26 & Concat(L10, L25) & - & $64 \times 64 \times 128$ \\
27 & Conv + ReLU & $3 \times 3$ & $64 \times 64 \times 64$ \\
28 & Conv + ReLU & $3 \times 3$ & $64 \times 64 \times 64$ \\

29 & Upsampling & $2 \times 2$ & $128 \times 128 \times 64$ \\
30 & Concat(L5, L29) & - & $128 \times 128 \times 128$ \\
31 & Conv + ReLU & $3 \times 3$ & $128 \times 128 \times 64$ \\
32 & Conv + ReLU & $3 \times 3$ & $128 \times 128 \times 64$ \\

33 & Upsampling & $2 \times 2$ & $256 \times 256 \times 64$ \\
34 & Conv + ReLU & $3 \times 3$ & $256 \times 256 \times 64$ \\
35 & Conv + ReLU & $3 \times 3$ & $256 \times 256 \times 64$ \\
\midrule
36 & Conv + Sigmoid & $1 \times 1$ & $256 \times 256 \times 1$ \\

\midrule
 & Output: L36 ($\mbs$) & - & - \\
\bottomrule
\end{tabular}
}
\vspace{-1em}
\end{table}

\begin{table}[h]   
\caption{Architecture of encoder $f^{\text{E}}$ and keypoint head $f^{\text{H}}$ in fine-tuning stage. We use the architecture of~\cite{vaswani2017attention} as Transformer block.}
\vspace{0.5em}
\label{tab:finetune_E_H}
\centering
\scriptsize
\resizebox{\linewidth}{!}{
\begin{tabular}{c c c}
\toprule
Layer & Operation & Dimensionality\\
\midrule
& Input: rgb image $\mbx$ & $256 \times 256 \times 3$\\
\midrule
1 & Patch embedding $(\mathbf{x}^{p})$ & $256 \times 768$ \\
\midrule
2 & Class token $x^\text{cls}$ & $1 \times 768$ \\
\midrule
3 & Concat(L1, L2) & $257 \times 768$ \\ 
4 & Transformer block & $257 \times 768$ \\
5 & Transformer block & $257 \times 768$ \\
6 & Transformer block & $257 \times 768$ \\
7 & Transformer block & $257 \times 768$ \\
8 & Transformer block & $257 \times 768$ \\
9 & Transformer block & $257 \times 768$ \\
10 & Transformer block & $257 \times 768$ \\
11 & Transformer block & $257 \times 768$ \\
12 & Transformer block & $257 \times 768$ \\
13 & Transformer block & $257 \times 768$ \\
14 & Transformer block & $257 \times 768$ \\
15 & Transformer block $(\mbF_\text{img})$ & $257 \times 768$ \\
\midrule
16 & Exclude & $256 \times 768$ \\
17 & Reshape & $16 \times 16 \times 768$ \\
18 & Deconv + ReLU + BatchNorm & $32 \times 32 \times 256$\\
19 & Deconv + ReLU + BatchNorm & $64 \times 64 \times 256$\\
20 & Conv & $64 \times 64 \times 64K$\\
21 & Reshape & $K \times 64 \times 64 \times 64$ \\
\midrule
& Output: L21 ($\hat{\mbH}$) & - \\
\bottomrule
\end{tabular}
}
\vspace{-1em}
\end{table}

\begin{table}[h]   
\caption{Architecture of attention network $f^{\text{A}}$ in fine-tuning stage. We use the architecture of~\cite{vaswani2017attention} as Transformer block.}
\vspace{0.5em}
\label{tab:finetune_a}
\centering
\scriptsize
\resizebox{\linewidth}{!}{
\begin{tabular}{c c c}
\toprule
Layer & Operation & Dimensionality\\
\midrule
& Input: rgb image $\mbx$ & $256 \times 256 \times 3$\\
\midrule
1 & Patch embedding $(\mathbf{x}^{p})$ & $256 \times 768$ \\
\midrule
2 & Class token $x^\text{cls}$ & $1 \times 768$ \\
\midrule
3 & Concat(L1, L2) & $257 \times 768$ \\ 
4 & Transformer block ($\mba^\text{1}_\text{cls}$) & $1 \times 256$ \\
5 & Transformer block ($\mba^\text{2}_\text{cls}$) & $1 \times 256$ \\
6 & Transformer block ($\mba^\text{3}_\text{cls}$) & $1 \times 256$ \\
7 & Transformer block ($\mba^\text{4}_\text{cls}$) & $1 \times 256$ \\
8 & Transformer block ($\mba^\text{5}_\text{cls}$) & $1 \times 256$ \\
9 & Transformer block ($\mba^\text{6}_\text{cls}$) & $1 \times 256$ \\
10 & Transformer block ($\mba^\text{7}_\text{cls}$) & $1 \times 256$ \\
11 & Transformer block ($\mba^\text{8}_\text{cls}$) & $1 \times 256$ \\
12 & Transformer block ($\mba^\text{9}_\text{cls}$) & $1 \times 256$ \\
13 & Transformer block ($\mba^\text{10}_\text{cls}$) & $1 \times 256$ \\
14 & Transformer block ($\mba^\text{11}_\text{cls}$) & $1 \times 256$ \\
15 & Transformer block ($\mba^\text{12}_\text{cls}$) & $1 \times 256$ \\
16 & Concat(L4:L15) & $12 \times 256$ \\
\midrule
 & Output: L16 $(\bar{\mba}^\text{A}_\text{cls})$ & - \\
\bottomrule
\end{tabular}
}
\vspace{-1em}
\end{table}

\section{Implementation details}
We used the Pytorch library for our implementation and 4 RTX 3090 GPUs. We used the ViT-Base~\cite{dosovitskiy2020image} as the architecture of encoder $f^\text{E}$ and U-Net~\cite{ronneberger2015u} as the segmentation network $f^\text{S}$. We also used the AdamW~\cite{loshchilov2017decoupled} optimizer with different learning rate, 2.4e-3 and 5e-4 for pre-training and fine-tuning, respectively. We used the RGB images of 256 $\times$ 256 input. For data augmentation, we employed the random rotation and translation, and resize followed by color jitter. In pre-training stage, we set the batch-size, weight decay and total epoch as 4,096, 0.05 and 800, respectively. We used the cosine learning rate strategy~\cite{loshchilov2016sgdr} and warm-up~\cite{goyal2017accurate}. The hyper-parameters of pre-training stage $r$ and $\alpha$ were experimentally set to 0.75 and 4. In fine-tuning stage, we set the batch-size to 128. For hand pose estimation, we used 50 and 100 as the total epoch and $\lambda_\text{attn}$, respectively. Also, we set them to 100 and 10 for human pose estimation.  

\begin{table*}[!t]
\centering
\renewcommand{\arraystretch}{1.4}
\caption{Ablation studies of the hyper-parameters, masking ratio $r$ and $\alpha$. All experiments are conducted on both tasks. The best is bold-faced.}
\vspace{0.2cm}
\begin{tabular}{cc|cc|cc|cc}
\hline
\multicolumn{2}{c|}{\multirow{2}{*}{\textbf{Method}}} & EPE($\downarrow$) & MPJPE($\downarrow$) & \multicolumn{2}{c|}{\multirow{2}{*}{\textbf{Method}}} & EPE($\downarrow$) & MPJPE($\downarrow$)\\
 & & (Hand) & (Human) & & & (Hand) & (Human) \\ \hline
\multicolumn{1}{c|}{\multirow{3}{*}{$\alpha$ = 4}} & $r$ = 0.5 & 32.1 & 104.1 & \multicolumn{1}{c|}{\multirow{3}{*}{$r$ = 0.75}} & $\alpha$ = 2 & 31.1 & 98.6 \\ 
\multicolumn{1}{c|}{} & $r$ = 0.75 & \textbf{28.0} & \textbf{93.6} & \multicolumn{1}{c|}{} & $\alpha$ = 4 & \textbf{28.0} & \textbf{93.6} \\ 
\multicolumn{1}{c|}{} & $r$ = 0.95 & 29.4 & 105.9 & \multicolumn{1}{c|}{} & $\alpha$ = 8 & 29.8 & 99.0 \\ \hline
\end{tabular}
\label{tab:ablation_hyperparam}
\end{table*}

\section{Datasets summary.}
\subsection{3D hand pose estimation.}
\noindent \textbf{FreiHAND}~\cite{zimmermann2019freihand} is the large-scale 3D hand pose dataset, which contains 130K samples with MANO~\cite{MANO:SIGGRAPHASIA:2017} parameters. It provides the train split created indoor environments and test split created both indoor and outdoor. We follow the splits as~\cite{zhang2021learning}.

\noindent \textbf{STB}~\cite{mueller2017stb} is real dataset of 3D hand pose estimation and contains left hand-only images captured in 6 indoor environments. We follow the train/test splits of Zimmerman~\etal~\cite{zimmermann2019freihand}.

\noindent \textbf{RHD}~\cite{zimmerman_iccv2017} is a synthetic dataset generated by Blender\footnote{https://www.blender.org} software. They sampled images from Flickr\footnote{https://www.flickr.com} and used them as the background of samples. It provides 43K samples with annotations for hand keypoints and segmentation masks. We follow the train/test splits of Zimmerman \etal~\cite{zimmerman_iccv2017}. 

\noindent \textbf{PANoptic (PAN)}~\cite{joo2018total} contains samples which actors perform the certain task or interact with each other in the studio. It was recorded using a multiview capture setup with 10 RGB-D sensors, 480 VGA and 31 HD cameras. We follow the train/test splits as~\cite{zimmermann2019freihand}.

\noindent \textbf{GANerated (GAN)}~\cite{mueller2018ganerated} is a synthetic dataset using the method of image-to-image translation CycleGAN~\cite{zhu2017unpaired} to make more realistic samples. It includes 330K samples which are hand-only or hand-object scenarios in random background. We randomly split them into train/test splits as~\cite{zimmermann2019freihand}.

\subsection{3D human pose estimation.}

\noindent \textbf{Human3.6M (H3.6M)}~\cite{ionescu2013human3} provides 3.6M indoor images containing 11 actors and 4 camera views. We follow the standard protocol using the subjects 1, 5, 6, 7 and 8 for training.   

\noindent \textbf{3DPW}~\cite{von2018recovering} contains outdoor 60 videos using the mobile phone and 17 IMU sensors for collecting the data. We follow the train/test splits of dataset.

\noindent \textbf{MPI-INF-3DHP (3DHP)}~\cite{mehta2017monocular} provides samples in both indoor and outdoor scenes containing 8 actors and 14 camera views. The evaluation is performed on the test split of dataset.

\noindent \textbf{SURREAL}~\cite{varol2017learning} is the synthetic dataset of 3D human pose estimation task. It provides the 63K videos generated by rendering the human on the random background. As~\cite{zhang2021learning}, we perform the evaluation on the validation split of dataset.

\subsection{Unconstrained dataset.}
\noindent \textbf{ImageNet}~\cite{krizhevsky2017imagenet} is the large-scale dataset for image classification task and includes 1.4M samples categorized into 1,000 classes. We only used the validation split of dataset for training.  

\section{Additional results.}
\begin{figure*}[!t]
\centering
\includegraphics[width=0.95\linewidth]{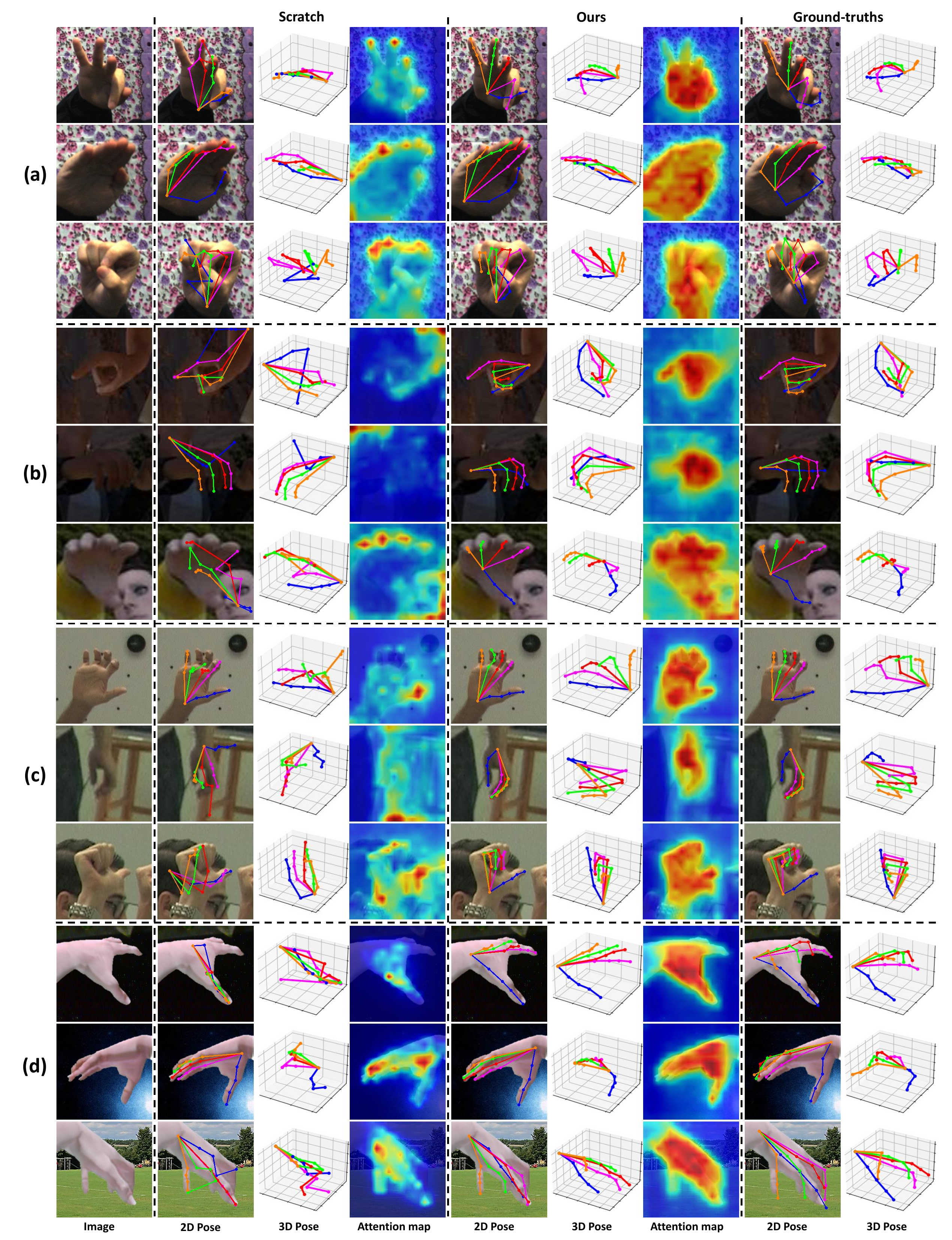}
\caption{Qualitative comparisons of scratch and ours on (a) STB~\cite{mueller2017stb}, (b) RHD~\cite{zimmerman_iccv2017}, (c) PANoptic (PAN)~\cite{joo2018total}, and (d) GANerated (GAN)~\cite{mueller2018ganerated} datasets. We involve the results of 2D pose, 3D pose and attention map for scratch and ours, respectively.}
\label{fig:hand_qualitative}
\end{figure*}

\begin{figure*}[!t]
\centering
\includegraphics[width=1\linewidth]{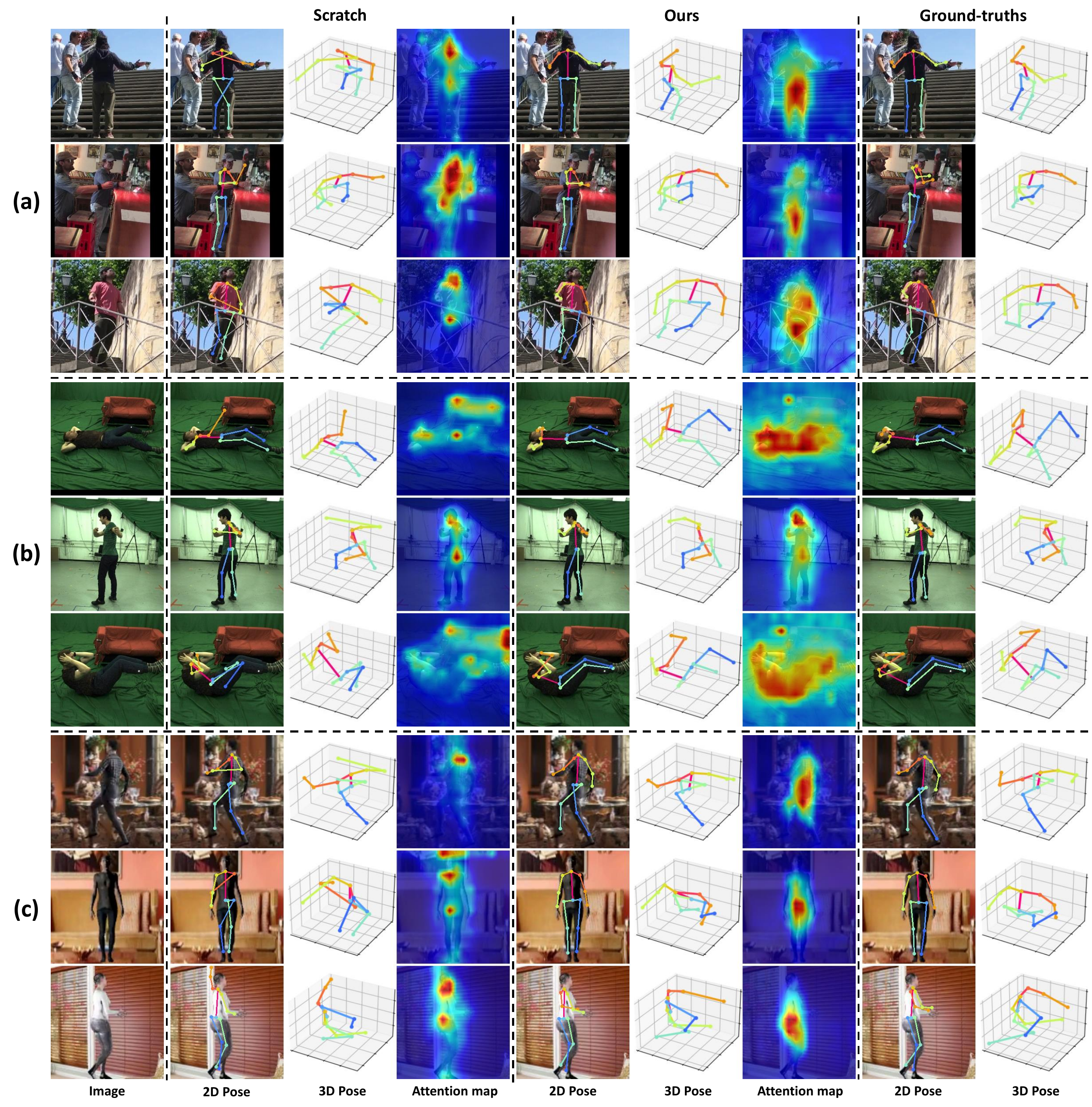}
\caption{Qualitative comparisons of scratch and ours on (a) 3DPW~\cite{von2018recovering}, (b) MPI-INF-3DHP (3DHP)~\cite{mehta2017monocular}, and (c) SURREAL~\cite{varol2017learning} datasets. We involve the results of 2D pose, 3D pose and attention map for scratch and ours, respectively.}
\label{fig:human_qualitative}
\end{figure*}

In Fig~\ref{fig:hand_qualitative} and~\ref{fig:human_qualitative}, we show the more qualitative comparisons with scratch on hand and human pose estimation task. The qualitative results contain the visualization for 2D pose, 3D pose and attention map. In hand pose estimation task, we evaluated on STB~\cite{mueller2017stb}, RHD~\cite{zimmerman_iccv2017}, PANoptic (PAN)~\cite{joo2018total} and GANerated (GAN)~\cite{mueller2018ganerated}. In human pose estimation task, we evaluated on 3DPW~\cite{von2018recovering}, MPI-INF-3DHP (3DHP)~\cite{mehta2017monocular} and SURREAL~\cite{varol2017learning}.   

\section{Ablation study on hyper-parameters.}
We conducted the experiments of hyper-parameters, masking ratio $r$ and $\alpha$ on both tasks. We used FreiHAND and RHD datasets as source and target domain dataset in hand pose estimation task. In human pose estimation task, Human3.6M and MPI-INF-3DHP (3DHP) dataset are used as source and target domain dataset, respectively. We used $\alpha = 4$ in the experiment of masking ratio $r$ and $r$ = 0.75 in the experiment of $\alpha$. Table.~\ref{tab:ablation_hyperparam} shows the quantitative result of experiments. We could observe that $r$ = 0.75 and $\alpha$ = 4 in each experiment show the best performance in both tasks.

\begin{algorithm*}[ht]
\setstretch{1.35}
\SetAlgoLined
\textbf{Pre-training stage.} \\
\textbf{Input: Image $\mbx = [\mbx_s, \mbx_t]$, Unconstrained image $\mbx_c$, Segmentation mask $\mbs = [\mbs_s, \mbs_t]$} \\
\textbf{Output: Reconstructed image $\hat{\mbx}$} \\
\For{$t=1,\ldots,T$}{

    

    Augment the image $\mbx$ with $\mbx_c$ and $\mbs$. \\
    
    Obtain the unmaksed tokens $\mbx^\text{p}_\text{m}$ from $\mbx$ and concatenate them with the class token $x^\text{cls}$. \\
    
    Feed them into the transformer blocks of $f^\text{E}$ and obtain the feature $\mbF_\text{MAE}$. \\
    
    Concatenate the feature $\mbF_\text{MAE}$ with mask tokens $\mbM$ and arrange them according to the token order of image. \\
    
    Feed them into $f^\text{D}$ and obtain the reconstructed image $\hat{\mbx}$. \\

    Transform the segmentation mask $\mbs$ into the weight $W$. \\
    
    Calculate gradient $\nabla L_{\text{WMAE}}$ and update $f^\text{E}$ and $f^\text{D}$.\\
}
\hrulefill \\

\textbf{Fine-tuning stage.} \\
\textbf{Input: Image $\mbx = [\mbx_s, \mbx_t]$} \\
\textbf{Output: heatmap $\hat{\mbH}$, attention map $\bar{\mba}^\text{E}_\text{cls}$, $\bar{\mba}^\text{A}_\text{cls}$} \\
\textbf{Initialization: pre-trained $f^\text{E}$} \\

\For{$t=1,\ldots,T'$}{

    Feed the source image $\mbx_s$ into $f^\text{E}$ and obtain the feature $\mbF_\text{img}$. \\
    
    Feed the feature $\mbF_\text{img}$ into $f^\text{H}$ and obtain the heatmap $\hat{\mbH}$. \\ 

    Calculate gradient $\nabla L_{\text{kpt}}$ and update $f^\text{E}$ and $f^\text{D}$.\\

    Feed the target image $\mbx_t$ into $f^\text{E}$ and $f^\text{A}$ and obtain the attention map $\bar{\mba}^\text{E}_\text{cls}$ and $\bar{\mba}^\text{A}_\text{cls}$.

    Calculate gradient $\nabla L_{\text{attn}}$ and update $f^\text{E}$.\\
}

\caption{The summary of the entire training procedure}
\label{a:training_procedure}
\end{algorithm*}

\end{document}